\documentclass[twoside,leqno,twocolumn]{article}
\usepackage{ltexpprt}

\usepackage{graphicx} 
\usepackage{subfigure} 
\usepackage{algorithm}
\usepackage{algorithmic}
\usepackage{amsfonts}
\usepackage{amssymb}
\usepackage{amsmath}
\usepackage{centernot}
\usepackage{blkarray}
\usepackage{url}

\begin{document}

\title{Multi-Region Neural Representation:\\ A novel model for decoding visual stimuli in human brains}
\author{Muhammad Yousefnezhad\thanks{Department of Computer Science and Technology, Nanjing University of Aeronautics and Astronautics Nanjing, China, email: myousefnezhad@nuaa.edu.cn.} \\
	\and
	Daoqiang Zhang\thanks{Department of Computer Science and Technology, Nanjing University of Aeronautics and Astronautics Nanjing, China, email: dqzhang@nuaa.edu.cn.}}
\date{}

\maketitle


\begin{abstract} \small\baselineskip=9pt
Multivariate Pattern (MVP) classification holds enormous potential for decoding visual stimuli in the human brain by employing task-based fMRI data sets. There is a wide range of challenges in the MVP techniques, i.e. decreasing noise and sparsity, defining effective regions of interest (ROIs), visualizing results, and the cost of brain studies. In overcoming these challenges, this paper proposes a novel model of neural representation, which can automatically detect the active regions for each visual stimulus and then utilize these anatomical regions for visualizing and analyzing the functional activities. Therefore, this model provides an opportunity for neuroscientists to ask this question: what is the effect of a stimulus on each of the detected regions instead of just study the fluctuation of voxels in the manually selected ROIs. Moreover, our method introduces analyzing snapshots of brain image for decreasing sparsity rather than using the whole of fMRI time series. Further, a new Gaussian smoothing method is proposed for removing noise of voxels in the level of ROIs. The proposed method enables us to combine different fMRI data sets for reducing the cost of brain studies. Experimental studies on 4 visual categories (words, consonants, objects and nonsense photos) confirm that the proposed method achieves superior performance to state-of-the-art methods.
\end{abstract}

\section{Introduction}
A universal unanswered question in neuroscience is how the human brain activities can be mapped to the different brain tasks? As one of the main techniques in task-based functional Magnetic Resonance Imaging (fMRI) analysis, Multivariate Pattern (MVP) is a conjunction between neuroscience and computer science, which can extract and decode brain patterns by applying the classification methods \cite{anderson10,lorbert12}. Indeed, it can predict patterns of neural activities associated with different cognitive states \cite{mohr15,mcmenamin15} and also can define decision surfaces to distinguish different stimuli for decoding the brain and understanding how it works \cite{haxby14,hanson04}. Analyzing the patterns of visual objects is one of the most interesting topics in MVP classification, which can enable us to understand how brain stores and processes the visual stimuli. It can be used to find novel treatments for mental diseases or even to create a new generation of the user interface.

Technically, MVP classification is really a challenging problem. Firstly, most of the fMRI data sets are noisy and sparse, which can decrease the performance of MVP methods \cite{chen15}. The next challenge is defining the regions of interest (ROIs) \cite{mcmenamin15}. As mentioned before, fMRI techniques allow us to study what information are represented in the different regions. So, it is really important to know what are the effects of different stimuli on the brain regions, especially in complex tasks (doing some simple tasks at the same time such as watching photos and tapping keys). On the one hand, most of the previous studies manually selected the ROIs. On the other hand, defining wrong ROIs can significantly decrease the performance of MVP methods \cite{mohr15,mcmenamin15}. Another challenge is the cost of brain studies. Combining different homogeneous fMRI data sets can be considered as a solution for this problem but data must be normalized in a standard space. The procedure of normalization can increase the time and space complexities and decrease the robustness of MVP techniques, especially in voxel-based methods \cite{haxby14}. The last challenge is visualization. As a machine learning technique, MVP represents the numerical results in the voxel-level, network connections, etc. Sometimes, it is so hard for neuroscientists to find a relation between the generated results and the cognitive states.

The contributions of the paper are four fold: firstly, the proposed method estimates and analyzes a snapshot of brain image for each stimulus based on the level of using oxygen in the brain instead of analyzing whole of fMRI time series. Indeed, employing these snapshots can dramatically decrease the sparsity. Secondly, our methods can automatically detect active regions for each stimulus and dynamically define ROIs for each data set. Further, it develops a novel model of neural representation for analyzing and visualizing functional activities in the form of anatomical regions. This model can provide a compact and informative representation of neural activities for neuroscientists to understand: what is the effect of a stimulus on each of the automatically detected regions instead of just study the fluctuation of a group of voxels in the manually selected ROIs. The next contribution is a new Gaussian smoothing method for removing noise of voxels in the level of anatomical regions. Lastly, this paper employs the L1-regularization Support Vector Machine (SVM) \cite{bradley98} method for creating binary classification at the ROIs level and then combine these classifiers by using the Bagging algorithm \cite{breiman96,murphy12} for generating the MVP model. 
\section{Related Works}
As the most prevalent techniques in the human brain decoding, MVP methods can predict patterns of neural activities. Since spatial resolution and within-area patterns of response in fMRI can provide an informative representation of stimulus distinctions, most of previous MVP studies for decoding the human brain focused on task-based fMRI data sets \cite{haxby14}. They used these data sets for generating different forms of neural representation, include usually voxels (volume elements in brain images), nodes on the cortical surface, the average signal for an area, a principal or independent component, or a measure of functional connectivity between a pair of locations \cite{haxby14,hanson04,haxby01,osher15}. Previous studies demonstrated that MVP classification can also distinguish many other brain states such as recognizing visual \cite{haxby14,hanson04,haxby01}, or auditory stimuli \cite{formisano08}.

Pioneer studies just focused on the special regions of the human brain, such as the Fusiform Face Area (FFA) or Parahippocampal Place Area (PPA) \cite{haxby01}. Haxby et al. showed that different visual stimuli, i.e. human faces, animals, etc., represent different responses in the brain \cite{haxby14,haxby12}. Hanson et al. developed combinatorial codes in the ventral temporal lobe for object recognition \cite{hanson04}. Norman et al. argued for using SVM and Gaussian Naive Bayes classifiers \cite{norman06}. Anderson and Oates studied the chance of applying non-linear Artificial Neural Network (ANN) on brain responses \cite{anderson10}. 

There is great potential for employing sparse methods for brain decoding problems \cite{yamashita08,ryali10}. Carroll et al. employed the Elastic Net \cite{zou05} for prediction and interpretation of distributed neural activity with sparse models \cite{carroll09}. Richiardi et al. extracted the characteristic connectivity signatures of different brain states to perform classification \cite{richiardi11}. Varoquaux et al. proposed a small-sample brain mapping by using sparse recovery on spatially correlated designs with randomization and clustering. Their method is applied on small sets of brain patterns for distinguishing different categories based on a one-versus-one strategy \cite{varoquaux12}. McMenamin et al. studied subsystems underlie abstract-category (AC) recognition and priming of objects (e.g., cat, piano) and specific-exemplar (SE) recognition and priming of objects (e.g., a calico cat, a different calico cat, a grand piano, etc.). Technically, they applied SVM on manually selected ROIs in the human brain for generating the visual stimuli predictors \cite{mcmenamin15}. Mohr et al. compared four different classification methods, i.e. L1/2 regularized SVM \cite{bradley98,cortes95}, the Elastic Net, and the Graph Net \cite{grosenick13}, for predicting different responses in the human brain. They show that L1-regularization can improve classification performance while simultaneously providing highly specific and interpretable discriminative activation patterns \cite{mohr15}. Osher et al. proposed a network (graph) based approach by using anatomical regions of the human brain for representing and classifying the different visual stimuli responses (faces, objects, bodies, scenes) \cite{osher15}. 
\section{The Proposed Method}
The fMRI techniques visualize the neural activities by measuring the level of oxygenation or deoxygenation in the human brain, which is called Blood Oxygen Level Dependent (BOLD) signals. Technically, these signals can be represented as time series for each subject. Most of the MVP techniques directly analyze these noisy and sparse time series for understanding which patterns are demonstrated for different stimuli. 

The main idea of our proposed method is so simple. Instead of analyzing whole of the time series, the proposed method estimates and analyzes a snapshot of brain image for each stimulus when the level of using oxygen is maximized. As a result, this method can automatically decrease the sparsity of brain image. The proposed method is applied in three stages: firstly, snapshots of brain image are selected by finding local maximums in the smoothed version of the design matrix. Then, features are generated in three steps, including normalizing to standard space, segmenting the snapshots in the form of automatically detected anatomical regions, and removing noise by Gaussian smoothing in the level of ROIs. Finally, decision surfaces \cite{haxby14} are generated by utilizing the bagging method on binary classifiers, which are created by applying L1-regularized SVM on each of neural activities in the level of ROIs.
\begin{figure*}[!t]
	\begin{center}
		\begin{minipage}{0.48\linewidth}
			\includegraphics[width=0.98\textwidth,height=0.48\linewidth]{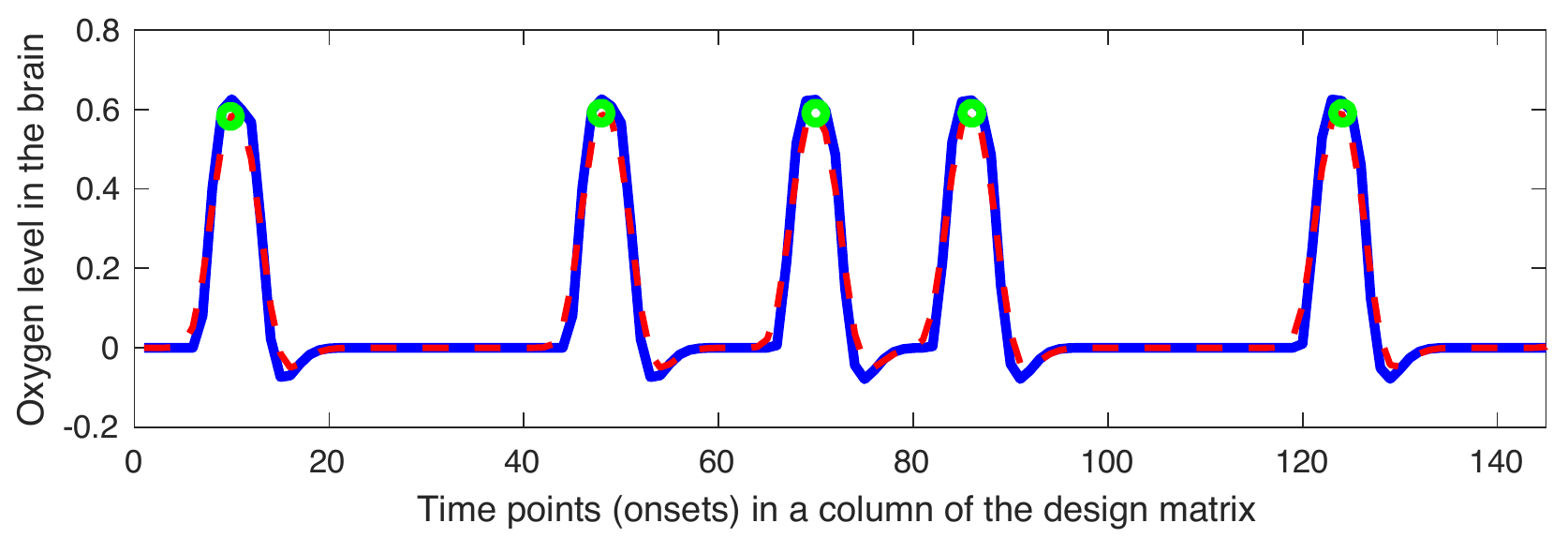}\\
			\centering (a) Design matrix in the block-design experiment
		\end{minipage}
		\begin{minipage}{0.48\linewidth}
			\includegraphics[width=0.98\textwidth,height=0.48\linewidth]{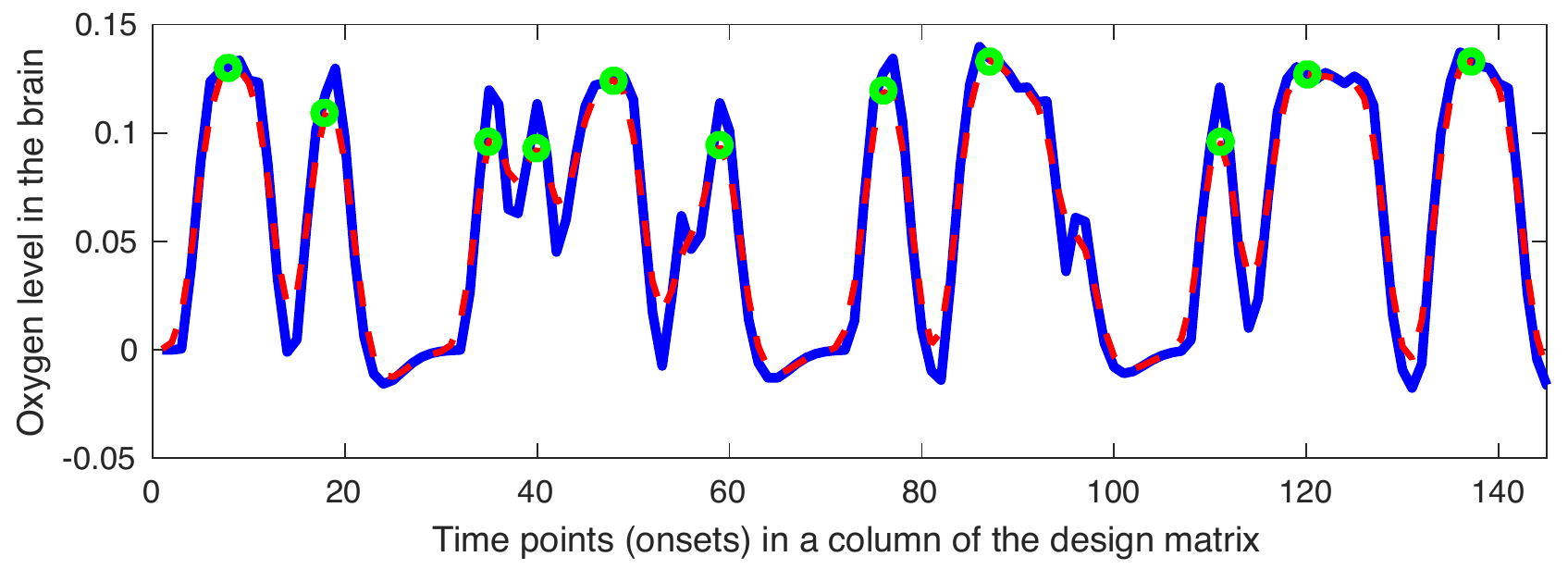}\\
			\centering (b) Design matrix in the event-related experiment
		\end{minipage}		
		\caption{Two examples of smoothed version of the design matrix. The blue lines show the original convolution ($\mathbf{d}_i = \mathbf{S}_i*\mathbf{H}$), the red dashed lines depict the smooth versions ($\phi_i = (\mathbf{S}_i*\mathbf{H})*\mathbf{G}$), and the green circles illustrate the locations ($\mathbf{S}_i^*$) of  the detected snapshots ($\mathbf{\widehat{\Psi}}$).}
		\label{SmoothedDM}
	\end{center}
	\vskip -0.3in
\end{figure*}
\begin{algorithm}[!t]
	\caption{The Snapshots Selection Algorithm}
	\label{alg:SnapshotSelection}
	\begin{algorithmic}
		\STATE {\bfseries Input:} fMRI time series $\mathbf{F}$, time points (onsets) $\mathbf{S}$, \\
		\qquad\qquad HRF signal $\mathbf{H}$, , Gaussian Parameter $\sigma_G$:\\
		\STATE {\bfseries Output:} Snapshots $\mathbf{\Psi}$, the sets of correlations $\widehat{\beta}$:\\
		\STATE {\bfseries Method:}\\
		\quad1. Generating the design matrix $\mathbf{D} = \mathbf{S}*\mathbf{H}$.\\
		\quad2. Defining $\mathbf{F} = \mathbf{D}\widehat{\beta}+\varepsilon$.\\
		\quad3. Calculating $\widehat{\beta}$ by using (\ref{BetaEq}).\\
		\quad4. Generating Gaussian kernel by (\ref{GaussianKernel}).\\
		\quad5. Smoothing the design matrix by (\ref{SmoothingDM}).\\
		\quad6. Finding locations of the snapshots by (\ref{SnapshotLocations}).\\
		\quad7. Calculating snapshots $\mathbf{\widehat{\Psi}}$ by using (\ref{Snapshots}).\\
	\end{algorithmic}
\end{algorithm}
\subsection{Snapshots Selection}
fMRI time series collected from a subject can be denoted by $\mathbf{F} \in \mathbb{R}^{t\times m}$, where $t$ is the number of time samples, and $m$ denotes the number of voxels. Same as previous studies \cite{anderson10,mcmenamin15,mohr15,haxby01},  $\mathbf{F}$ can be formulated by a linear model as follows: 
\begin{equation}
	\mathbf{F} = \mathbf{D}(\mathbf{\widehat{\beta}})^\intercal + \mathbf{\varepsilon} 
\end{equation}
where $\mathbf{D} \in \mathbb{R}^{t\times p}$ denotes the design matrix, $\varepsilon$ is the noise (error of estimation),  $\widehat{\beta} \in \mathbb{R}^{m\times p}$ denotes the sets of correlations (estimated regressors) between voxels. The design matrix can be denoted by $\mathbf{D} = \{\mathbf{d}_{1}, \mathbf{d}_{2}, \dots, \mathbf{d}_{i}, \dots, \mathbf{d}_{p}\}$, and the sets of correlations can be defined by $\widehat{\beta}=\{\widehat{\beta}_1, \widehat{\beta}_2, \dots, \widehat{\beta}_i, \dots, \widehat{\beta}_p \}$. Here, $\mathbf{d}_{i} \in \mathbb{R}^t$ and $\widehat{\beta}_i \in \mathbb{R}^m$ are the column of design matrix and the set of correlations for $i-th$ category, respectively. $p$ is also the number of all categories in the experiment $\mathbf{F}$. In fact, each category (independent tasks) contains a set of homogeneous visual stimuli. In addition, the nonzero voxels in $\widehat{\beta}_i$ represents the location of all active voxels for the $i-th$ category \cite{friston03}. As an example, imagine during a unique session for recognizing visual stimuli, if a subject watches 4 photos of cats and 3 photos of houses, then the design matrix contains two columns; and there are also two sets of correlations between voxels, i.e. one for watching cats and another for watching houses. Indeed, the final goal of this section is extracting 7 snapshots of the brain image for the 7 stimuli in this example. 

The design matrix can be classically calculated by convolution of time samples (or onsets: $\mathbf{S} = \{\mathbf{S}_1, \mathbf{S}_2, \dots, \mathbf{S}_i, \dots, \mathbf{S}_p\}$) and $\mathbf{H}$ as the Hemodynamic Response Function (HRF) signal, $\mathbf{d}_i = \mathbf{S}_i*\mathbf{H} \implies \mathbf{D} = \mathbf{S}*\mathbf{H}$ \cite{haxby01,friston03}. In addition, there is a wide range of solutions for estimating $\widehat{\beta}$ values. This paper uses the classical method Generalized Least Squares (GLS) \cite{friston03} for estimating the $\widehat{\beta}$ values where $\mathbf{\Sigma}$ is the covariance matrix of the noise ($Var(\mathbf{\varepsilon})=\mathbf{\Sigma}{\mathbf{\sigma}}^{2}\ne\mathbb{I}{\sigma}^{2}$): 
\begin{equation}
	\label{BetaEq}\widehat{\beta}=\big({({\mathbf{D}}^{\intercal}{\mathbf{\Sigma}}^{-1}\mathbf{D})}^{-1}{\mathbf{D}}^{\intercal}{\mathbf{\Sigma}}^{-1}\mathbf{F}\big)^\intercal
\end{equation}
Each local maximum in $\mathbf{d}_{i}$ represents a location where the level of using oxygen is so high. In other words, the stimulus happens in that location. Since $\mathbf{d}_{i}$ mostly contains small spikes (especially for event-related experiments), it cannot be directly used for finding these local maximums. Therefore, this paper employs a Gaussian kernel for smoothing the $\mathbf{d}_{i}$ signal.  Now, the interval $\mathbf{\widehat{G}}$ is defined as follows for generating the kernel:
\begin{equation}\label{eq:GHatInetval}
	\mathbf{\widehat{G}} = \bigg\{ \exp\left(\frac{-{\mathbf{\widehat{g}}}^{2}}{2{\sigma^2_G}}\right) \bigg| \text{ } \mathbf{\widehat{g}} \in \mathbb{Z} \text{ and } -2\lceil\sigma_G\rceil \leq \mathbf{\widehat{g}} \leq 2\lceil\sigma_G\rceil \bigg\}
\end{equation}
where $\sigma_G > 0$ denotes a positive real number; $\lceil . \rceil$ is the ceiling function; and $\mathbb{Z}$ denotes the set of integer numbers. Gaussian kernel is also defined by normalizing $\mathbf{\widehat{G}}$ as follows:
\begin{equation}
	\label{GaussianKernel}
	\mathbf{G} = \frac{\mathbf{\widehat{G}}}{\sum_{j}{\mathbf{\widehat{g}}_j}} 
\end{equation}
where $\sum_{j}{\mathbf{\widehat{g}}_j}$ is the sum of all elements in the interval $\mathbf{\widehat{G}}$. This paper defines the smoothed version of the design matrix by applying the convolution of the Gaussian kernel $\mathbf{G}$ and each column of the design matrix ($\mathbf{d}_{i}$) as follows:
\begin{equation}
	\label{SmoothingDM}
	\mathbf{\phi}_{i} = \mathbf{d}_{i}*\mathbf{G} = (\mathbf{S}_{i}*\mathbf{H})*\mathbf{G}
\end{equation}
\begin{equation}
	\mathbf{\Phi} = \{\mathbf{\phi}_{1}, \mathbf{\phi}_{2}, \dots, \mathbf{\phi}_{p}\}
\end{equation}
where $\phi_{i} = f\left(\mathbf{S}_{i},\mathbf{H},\mathbf{G}\right)$. Since the level of smoothness in $\mathbf{\Phi}$ is related to the positive value in \eqref{eq:GHatInetval}, $\sigma_G=1$  is heuristically defined to generate the optimum level of smoothness in the design matrix. The general assumption here is the $0 < \sigma_G < 1$ can create design matrix, which is sensitive to small spikes. Further, $\sigma_G > 1$ can rapidly increase the level of smoothness, and remove some weak local maximums, especially in the event-related fMRI data sets. Figure \ref{SmoothedDM} illustrates two examples of the smoothed columns in the design matrix. The local maximum points in the $\phi_{i}$ can be calculated as follows: 
\begin{multline}
	\mathbf{S}_{i}^{*} = \bigg\{ \underset{\mathbf{S}_{i}}\arg \quad \phi_{i} \text{ }\bigg| \text{ } \frac{\partial\phi_{i}}{\partial \mathbf{S}_{i}} = 0 \text { and } \frac{\partial^2\phi_{i}}{\partial {\mathbf{S}_{i}}\mathbf{S}_{i}} > 0 \bigg\}
\end{multline} 
where $\mathbf{S}_{i}^{*} \subset \mathbf{S}_{i}$ denotes the set of time points for all local maximums in $\mathbf{\phi}_{i}$. The sets of maximum points for all categories can be denoted as follows:
\begin{equation}
	\label{SnapshotLocations}
	\mathbf{S}^{*} = \{\mathbf{S}_{1}^{*}, \mathbf{S}_{2}^{*}, \dots, \mathbf{S}_{i}^{*}, \dots, \mathbf{S}_{p}^{*}\}
\end{equation}
As mentioned before, the fMRI time series can be also denoted by $\mathbf{F}^\intercal = \{\mathbf{f}_{1}^\intercal, \mathbf{f}_{2}^\intercal,\dots,\mathbf{f}_{j}^\intercal,\dots,\mathbf{f}_{t}^\intercal\}$, where $\mathbf{f}_{j}^\intercal \in \mathbb{R}^m$ is all voxels of fMRI data set in the $j-th$ time point. Now, the set of snapshots can be formulated as follows:
\begin{multline}
	\label{Snapshots}
	\mathbf{\widehat{\Psi}} = \{\mathbf{f}_{j}^\intercal \text{ } | \text{ } \mathbf{f}_{j}^\intercal \in \mathbf{F}^\intercal \text{ and } j \in \mathbf{S}^{*}\} = \\
	\{\widehat{\mathbf{\psi}_{1}}, \widehat{\mathbf{\psi}_{2}}, \dots, \widehat{\mathbf{\psi}_{k}}, \dots \widehat{\psi_{q}}\} \in \mathbb{R}^{m\times q}
\end{multline}
where $q$ is the number of snapshots in the brain image $\mathbf{F}$, and $\widehat{\psi_{k}} \in \mathbb{R}^m$ denotes the snapshot for $k-th$ stimulus. These selected snapshots are employed in next section for extracting features of the neural activities. Algorithm \ref{alg:SnapshotSelection} illustrates the whole of procedure for generating the snapshots from the time series $\mathbf{F}$.
\subsection{Feature Extraction}
In this paper, the feature extraction is applied in three steps, i.e. normalizing snapshots to standard space, segmenting the snapshots in the form of automatically detected regions, and removing noise by Gaussian smoothing in the level of ROIs. As mentioned before, normalizing brain image to the standard space can increase the time and space complexities and decrease the robustness of MVP techniques, especially in voxel-based methods \cite{haxby14}. On the one hand, most of the previous studies \cite{mohr15,mcmenamin15,hanson04,haxby01} preferred to use original data sets instead of the standard version because of the mentioned problem. On the other hand, this mapping can provide a normalized view for combing homogeneous data sets. As a result, it can significantly reduce the cost of brain studies and rapidly increase the chance of understanding how the brain works. Employing brain snapshots rather than analyzing whole of data can solve the normalization problem. 

Normalization can be formulated as a mapping problem.  Indeed, brain snapshots are mapped from $\mathbb{R}^m$ space  to the standard space $\mathbb{R}^n$ by using a transformation matrix for each snapshot. There is also another trick for improving the performance of this procedure. Since the set $\widehat{\beta}_i$ denotes the locations of all active voxels for the $i-th$ category, it represents the brain mask for that category and can be used for generating the transform matrix related to all snapshots belong to that category. For instance, in the example of the previous section, instead of calculating 7 transform matrices for 7 stimuli, we calculate 2 matrices, including one for the category of cats and the second one for the category of houses. This mapping can be denoted as follows:
\begin{equation}
	\mathbf{T}_i\text{:} \qquad\quad\widehat{\beta}_i \in \mathbb{R}^m \quad\to\quad \beta_i \in \mathbb{R}^n
\end{equation}
where $\mathbf{T}_i \in \mathbb{R}^{m \times n}$ denotes the transform matrix, ${\beta}_i = \big((\widehat{\beta}_i)^\intercal \mathbf{T}_i\big)^\intercal$ is the set of correlations in the standard space for $i-th$ category. This paper utilizes the FLIRT algorithm \cite{jenkinson02} for calculating the transform matrix, which minimizes the following objective function:
\begin{equation}
	\mathbf{T}_i = \arg\min(NMI(\widehat{\beta}_i,\mathbf{Ref}))
	\label{eq:ImageReg}
\end{equation}
where the function $NMI$ denotes the Normalized Mutual Information between two images \cite{jenkinson02}, and $\mathbf{Ref} \in \mathbb{R}^n$ is the reference image in the standard space. This image must contain the structures of the human brain, i.e. white matter, gray matter, and CSF. These structures can improve the performance of mapping between the brain mask in the selected snapshot and the general form of a standard brain. The performance of (\ref{eq:ImageReg}) will be analyzed in the supplementary materials\footnote{Supplementary Materials is available:\\\url{sourceforge.net/projects/myousefnezhad/files/MRNR/}}. In addition, the sets of correlations for all of categories in the standard space is denoted by $\beta = \{\beta_1, \beta_2, \dots, \beta_i, \dots, \beta_p\} \in \mathbb{R}^{n\times p}$, and the sets of transform matrices is defined by $\mathbf{T} = \{\mathbf{T}_1, \mathbf{T}_2, \dots, \mathbf{T}_i, \dots, \mathbf{T}_p \}$. Now, the $Select$ function is denoted as follows to find suitable transform matrix for each snapshot: 
\begin{equation}\label{SelectFunc}
\begin{split}
\big(\mathbf{T}^*_j, \beta^*_j\big) = Select(\widehat{\psi_j}, \mathbf{T}, \beta) = \{(\mathbf{T}_i, \beta_i) \text{ } | \qquad\quad\\ 
\text{ } \mathbf{T}_i \in \mathbf{T}\text{, } \beta_i \in \beta\text{ , }\widehat{\psi_j} \text{ is belonged to the $i-th$}\quad\\ 
\text{ category} \implies \widehat{\psi_j} \propto \beta_i \propto \mathbf{T}_i \}\quad
	\end{split}
\end{equation}
where $\mathbf{T}^*_j \in \mathbb{R}^{m \times n}$ and $\beta^*_j \in \mathbb{R}^{n}$ are the transform matrix and the set of correlations related to the $j-th$ snapshot, respectively. Based on (\ref{SelectFunc}), each normalized snapshot in the standard space is defined as follows:
\begin{equation}
	\begin{split}\label{eq:SnapshotStandardSpace}
		\mathbf{T}_j^*\text{:  } \widehat{\psi}_j \in \mathbb{R}^m \to \psi_j \in \mathbb{R}^n\implies 
		\psi_j = \bigg(\big(\widehat{\psi}_j\big)^\intercal \mathbf{T}^*_j \bigg)^\intercal
	\end{split}
\end{equation}
where $\psi_j \in \mathbb{R}^{n}$ is the $j-th$ snapshot in the standard space. Further, all snapshots in the standard space can be defined by $\mathbf{\Psi} = \{\psi_1, \psi_2, \dots, \psi_j, \dots, \psi_q \} \in \mathbb{R}^{n \times q}$. As mentioned before, nonzero values in the correlation sets depict the location of the active voxels. Based on (\ref{SelectFunc}), this paper uses these correlation sets as weights for each snapshot as follows:
\begin{equation}\label{NonZeroSnapshots}
	\mathbf{\Theta}_j = \mathbf{\psi}_j \circ \mathbf{\beta}_j^*
\end{equation}
where $\circ$ denotes Hadamard product, and $\mathbf{\Theta}_j \in \mathbb{R}^n$ is the $j-th$ modified snapshot, where the values of deactivated voxels (and also deactivated anatomical regions) are zero in this snapshot. As the final product of normalization procedure, the set of snapshots can be denoted by $\mathbf{\Theta} = \{\mathbf{\Theta}_1, \mathbf{\Theta}_2, \dots, \mathbf{\Theta}_j, \dots, \mathbf{\Theta}_q \}$. Further, each snapshot can be defined in the voxel level as follows, where $\mathbf{\theta}_j^k$ is the $k-th$ voxel of $j-th$ snapshot:
\begin{equation}
	\mathbf{\Theta}_j = \big[\theta_j^1, \theta_j^2, \dots, \theta_j^k, \dots, \theta_j^n \big]
\end{equation}
\begin{figure*}
	\begin{center}
		\begin{minipage}{0.48\linewidth}
			\includegraphics[width=0.98\textwidth,height=0.48\linewidth]{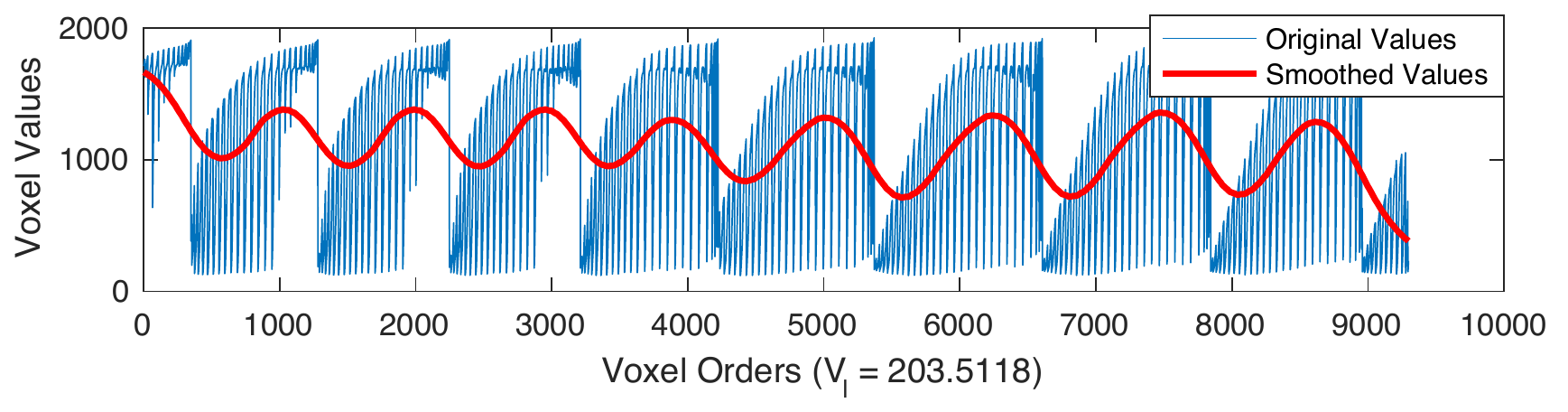}
		\end{minipage}
		\begin{minipage}{0.48\linewidth}
			\includegraphics[width=0.98\textwidth,height=0.48\linewidth]{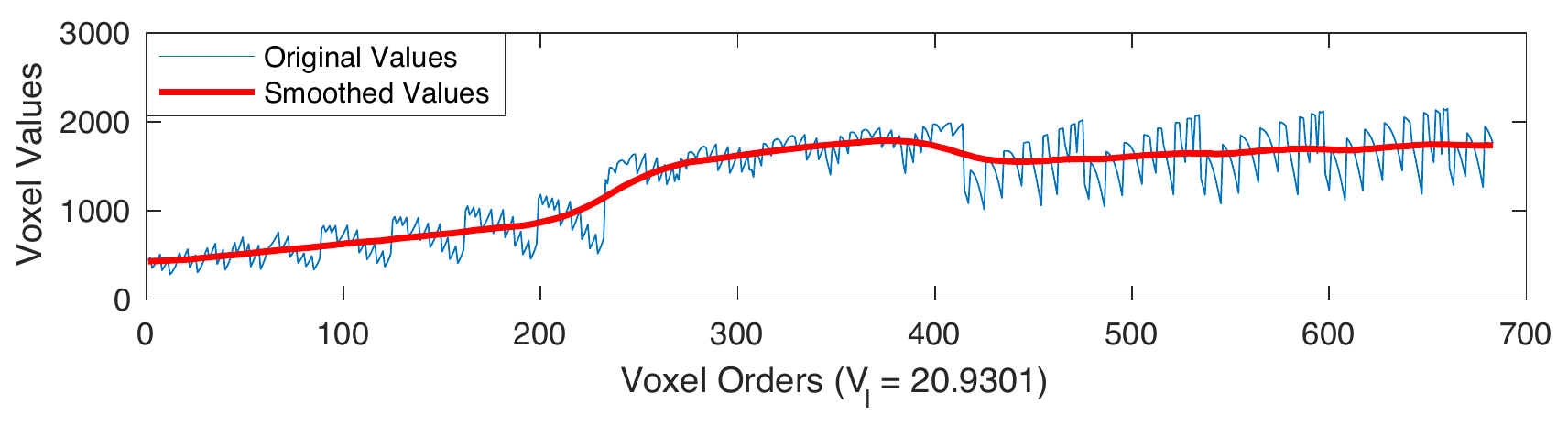}
		\end{minipage}		
		\caption{Two examples of smoothed anatomical regions ($\mathbf{X}_{(j,\ell)}$) in the voxel level. Blue lines are the original data, and red lines depict the smoothed values.}
		\label{fig:SmoothedAR}
	\end{center}
\end{figure*}
\begin{algorithm}[t]
	\caption{The Feature Extraction Algorithm}
	\label{alg:FeatureExtraction}
	\begin{algorithmic}
		\STATE {\bfseries Input:} Snapshots $\mathbf{\Psi}$, correlations $\widehat{\beta}$, $\mathbf{Ref}$ image, Atlas $\mathbf{A}$:\\
		\STATE {\bfseries Output:} Smoothed snapshots $\mathbf{X}$:\\
		\STATE {\bfseries Method:}\\
		1. For each $\widehat{\beta}_i$, calculate transform matrix by (\ref{eq:ImageReg}).\\ 
		2. Mapping $\widehat{\psi}_j$ to standard space by $\mathbf{T}^*_j$ and (\ref{eq:SnapshotStandardSpace}).\\
		3. Detecting active voxels for each snapshot by (\ref{NonZeroSnapshots}).\\
		4. Segmenting each snapshot by (\ref{eq:SnapshotSegmentation}).\\
		5. Finding active regions for each snapshot by (\ref{AutomaticallyDetectedActiveRegions}).\\
		6. Generating Gaussian kernel by (\ref{GaussianKernelForRegions}).\\
		7. Smoothing snapshots by (\ref{eq:SmoothedSnapshots}).
	\end{algorithmic}
\end{algorithm}
The next step is segmenting the snapshots in the form of automatically detected regions. Now, consider anatomical atlas $\mathbf{A} \in \mathbb{R}^n = \{\mathbf{A}_1, \mathbf{A}_2,\dots, \mathbf{A}_\ell, \dots, \mathbf{A}_L\}$, where ${\cap}_{\ell=1}^{L}\{\mathbf{A}_\ell\}=\emptyset$, ${\cup}_{\ell=1}^{L}\{\mathbf{A}_\ell\}=\mathbf{A}$, and $L$ is the number of all regions in the anatomical atlas. Here, $\mathbf{A}_\ell$ denotes the set of voxel locations in the snapshots for the $\ell-th$ anatomical region. A segmented snapshot based on the $\ell-th$ region can be denoted as follows: 
\begin{equation}
	\label{eq:SnapshotSegmentation}
	\mathbf{\Theta}_{(j,\ell)} = \{\theta_j^k \text{ } | \text{ } \theta_j^k \in \mathbf{\Theta}_j \text{ and } k \in \mathbf{A}_\ell \}
\end{equation}
where $\mathbf{\Theta}_{(j,\ell)} \subset \mathbf{\Theta}_j$ is the subset of voxels in the snapshot $\mathbf{\Theta}_j$, which these voxels are belonged to the the $\ell-th$ anatomical region. In addition, the sets of all anatomical regions in the $j-th$ snapshot can be defined by $\mathbf{\Theta}_j = \{\mathbf{\Theta}_{(j,1)}\cup \mathbf{\Theta}_{(j,2)}\cup \dots\cup \mathbf{\Theta}_{(j,\ell)}\cup \dots\cup \mathbf{\Theta}_{(j,L)}\} =  \big[\theta_j^1, \theta_j^2, \dots, \theta_j^k, \dots, \theta_j^n \big]$. The automatically detected active regions can be also defined as follows:
\begin{equation}
	\label{AutomaticallyDetectedActiveRegions}
	\mathbf{\Theta}_j^* = \bigg\{\mathbf{\Theta}_{(j,\ell)} | \mathbf{\Theta}_{(j,\ell)} \subset \mathbf{\Theta}_j \text { and } \sum_{\theta_{(j,\ell)}^{k} \in \mathbf{\Theta}_{(j,\ell)}} | \theta_{(j,\ell)}^{k} | \neq 0 \bigg\}
\end{equation}
where $\sum_{\theta_{(j,\ell)}^{k} \in \mathbf{\Theta}_{(j,\ell)}} | \theta_{(j,\ell)}^{k} | $ represents sum of all voxels in the $\mathbf{\Theta}_{(j,\ell)}$. Based on (\ref{AutomaticallyDetectedActiveRegions}), active regions in the $j-th$ snapshot can be defined as the regions with non-zero voxels because values of all deactivated voxels are changed to zero by using (\ref{NonZeroSnapshots}). The last step is removing noise by Gaussian smoothing in the level of ROIs. As the first step, a Gaussian kernel for each anatomical region can be defined as follows:
\begin{multline}
	\label{GaussianKernelForRegions}
	\qquad\qquad\qquad\sigma_\ell = \frac{N_\ell^2}{5N_\ell^2\log N_\ell}\\
	\mathbf{\widehat{V}_\ell} = \bigg\{ \exp\left(\frac{-{\mathbf{\widehat{v}}}^{2}}{2{\sigma_\ell}}\right) \bigg| \text{ } \mathbf{\widehat{v}} \in \mathbb{Z} \text{ and } -2\lceil\sigma_\ell\rceil \leq \mathbf{\widehat{v}} \leq 2\lceil\sigma_\ell\rceil \bigg\}\\
	\mathbf{V}_\ell = \frac{\widehat{\mathbf{V}_\ell}}{\sum_{j}{\mathbf{\widehat{v}}_j}} \qquad\qquad\qquad\qquad
\end{multline} 
where $N_\ell$ denotes the number of voxels in $\ell-th$ region, and $\sum_{j}{\mathbf{\widehat{v}}_j}$ is sum of all values in the interval $\mathbf{\widehat{V}}_\ell$. Indeed, the level of smoothness is related to $\sigma_\ell$, which is heuristically calculated for each region based on the number of voxels in that region. As the second step, the smoothed version of the $j-th$ snapshot can be defined as follows:
\begin{multline}\label{eq:SmoothedSnapshots}
	\qquad\forall \ell = L1 \dots L2 \to \mathbf{X}_{(j,\ell)} = \mathbf{\Theta}_{(j,\ell)} * \mathbf{V}_\ell,\\
	\mathbf{X}_{j} = \{\mathbf{X}_{(j,L1)}, \dots, \mathbf{X}_{(j,\ell)}, \dots \mathbf{X}_{(j,L2)}\} \qquad
\end{multline} 
where $\mathbf{\Theta}_{(j,\ell)} \in \mathbf{\Theta}_j^*$ is the $\ell-th$ active region of $j-th$ snapshot, and $*$ denotes the convolution between the active region and the Gaussian kernel related to that region. Further, $L1$ and $L2$ are the first and the last active regions in the snapshot, where $1 \leq L1 \leq L2 \leq L$. Figure \ref{fig:SmoothedAR} demonstrates two examples of smoothed anatomical regions in the voxel level. All smoothed snapshots can be defined by $\mathbf{X}=\{\mathbf{X}_1, \mathbf{X}_2, \dots, \mathbf{X}_j, \dots, \mathbf{X}_q \}$. Moreover, Algorithm \ref{alg:FeatureExtraction} shows the whole of procedure for extracting features.  
\subsection{Classification Method}
As a classical classification method, Support Vector Machine (SVM) \cite{bradley98,cortes95} decreases the operating risk and can find an optimized solution by maximizing the margin of error. As a result, it can mostly generate better performance in comparison with other methods, especially for binary classification problems. Therefore, SVM is generally used in the wide range of studies for creating predictive models \cite{mcmenamin15,mohr15,hanson04,haxby01}. The final goal of this section is employing the L1-regularization SVM \cite{bradley98} method for creating binary classification at the ROIs level, and then combining these classifiers by using the Bagging algorithm \cite{breiman96,murphy12} for generating the MVP final predictive model. 

As mentioned before, fMRI time series for a subject can be denoted by $\mathbf{F}$. Since fMRI experiment is mostly multi-subject, this paper denotes $\mathbf{F}_u, =1\text{:}U$ as fMRI time series (sessions) for all subjects, where $U$ is the number of subjects. In addition, $\tau = \sum_{u=1}^{U}q_u$ is defined as the number of all snapshots in a unique fMRI experiment. Here $q_u$ is the number of snapshots for $u-th$ subject. Further, the original ground truth (the title of stimuli such that cats, houses, etc.) for all snapshots is denoted by $\mathbf{Y} = \{y_1, y_2, \dots, y_j, \dots y_\tau\}$, where $y_j$ denotes the ground truth for $j-th$ snapshot. Since this paper uses a one-versus-all strategy, we can consider that $y_j \in \{-1, +1\}$. This paper applies following objective function on automatically detected active regions as the L1-regularization SVM method for creating binary classification in the level of ROIs \cite{mohr15,bradley98}:  
\begin{equation}\label{eq:L1SVM}
	\mathbf{\eta}_\ell\text{:} \quad \underset{\mathbf{W}_{\ell}}{\min}\text{\quad}C\sum_{j=1}^{\tau} \max(0,1-{y_j}{\mathbf{X}_{(j,\ell)}{\mathbf{W}_{(j,\ell)}}}) + \| \mathbf{W}_{\ell} \|_1
\end{equation}
where $C > 0$ is a real positive number, $\mathbf{X}_{(j,\ell)}$ and $y_j$ denote the voxel values of $\ell-th$ region and the class label of $j-th$ snapshot, respectively. Further, $\mathbf{W}_{\ell}=[ \mathbf{W}_{(1,\ell)}, \mathbf{W}_{(2,\ell)}, \dots, \mathbf{W}_{(j,\ell)}, \dots, \mathbf{W}_{(\tau,\ell)} ]$ is the generated weights for predicting MVP model based on the $\ell-th$ active region. The classifier for $\ell-th$ region is also denoted by $\mathbf{\eta}_\ell$, where all of these classifiers can be defined by $\mathbf{\eta} = \{\mathbf{\eta}_{L1}, \dots, \mathbf{\eta}_{\ell}, \dots \mathbf{\eta}_{L2}\}$. The final step in the proposed method is combining all classifiers ($\mathbf{\eta}$) by Bagging \cite{breiman96} algorithm for generating the MVP final predictive model. Indeed, Bagging method uses the average of predicted results in (\ref{eq:L1SVM}) for generating the final result ($\mathbf{\eta}_{final} = \sum_{\ell=L1}^{L2}\mathbf{\eta}_{\ell}$) \cite{breiman96,murphy12}. Algorithm \ref{alg:TheProposedMethod} shows the whole of procedure in the proposed method by using Leave-One-Out (LOO) cross-validation in the subject level.\\
\begin{table*}
	\caption{Accuracy of binary predictors  }
	\vskip -0.15in
	\label{tbl:BinaryAccuracy}
	\begin{center}
		\begin{small}
			\begin{tabular}{lcccccc}
				\hline
				Data Sets & SVM & Graph Net & Elastic Net &  L1-Reg. SVM & Osher et al. & Proposed method \\
				\hline
				DS105: Objects vs. Scrambles& 71.65$\pm$0.97 & 81.27$\pm$0.59 & 83.06$\pm$0.36 & 85.29$\pm$0.49 & 90.82$\pm$1.23 & \textbf{94.32$\pm$0.16}  \\
				DS107: Words vs. Others & 82.89$\pm$1.02 & 78.03$\pm$0.87 & 88.62$\pm$0.52 & 86.14$\pm$0.91 & 90.21$\pm$0.83 & \textbf{92.04$\pm$0.09}  \\
				DS107: Consonants vs. Others & 67.84$\pm$0.82 & 83.01$\pm$0.56 & 82.82$\pm$0.37 & 85.69$\pm$0.69 & 84.54$\pm$0.99 & \textbf{96.73$\pm$0.19}\\
				DS107: Objects vs. Others & 73.32$\pm$1.67 & 77.93$\pm$0.29 & 84.22$\pm$0.44 & 83.32$\pm$0.41 & \textbf{95.62$\pm$0.83} & {93.07$\pm$0.27}\\
				DS107: Scrambles vs. Others & 83.96$\pm$0.87 & 79.37$\pm$0.82 & 87.19$\pm$0.26 & 86.45$\pm$0.62 & 88.1$\pm$0.78 & \textbf{90.93$\pm$0.71}\\
				DS117: Faces vs. Scrambles & 81.25$\pm$1.03 & 85.19$\pm$0.56 & 85.46$\pm$0.29 & 86.61$\pm$0.61 & \textbf{96.81$\pm$0.79} & 96.31$\pm$0.92\\
				ALL: Faces vs. Others & 66.27$\pm$1.61 & 68.37$\pm$1.31 & 75.91$\pm$0.74 & 80.23$\pm$0.72 & 84.99$\pm$0.71 & \textbf{89.99$\pm$0.31}\\
				ALL: Objects vs. Others & 75.61$\pm$0.57 & 78.37$\pm$0.71 & 76.79$\pm$0.94 & 80.14$\pm$0.47 & 79.23$\pm$0.25 & \textbf{92.44$\pm$0.92}\\				
				ALL: Scrambles vs. Others & 81.92$\pm$0.71 & 81.08$\pm$1.23 & 84.18$\pm$0.42 & 88.23$\pm$0.81 & 90.5$\pm$0.73 & \textbf{95.39$\pm$0.18}\\				
				\hline
			\end{tabular}
		\end{small}
	\end{center}
	\vskip -0.25in
\end{table*}
\begin{table*}
	\caption{Area Under the ROC Curve (AUC) of binary predictors  }
	\vskip -0.15in
	\label{tbl:BinaryAUC}
	\begin{center}
		\begin{small}
			\begin{tabular}{lcccccc}
				\hline
				Data Sets & SVM & Graph Net & Elastic Net & L1-Reg. SVM & Osher et al. & Proposed method \\
				\hline
				DS105: Objects vs. Scrambles & 68.37$\pm$1.01 & 70.32$\pm$0.92 & 82.22$\pm$0.42 & 80.91$\pm$0.21 & 88.54$\pm$0.71 & \textbf{93.25$\pm$0.92}  \\
				DS107: Words vs. Others& 80.76$\pm$0.91 & 77.91$\pm$1.03 & 86.35$\pm$0.39 & 84.23$\pm$0.57 & 87.61$\pm$0.62 & \textbf{91.86$\pm$0.17}  \\
				DS107: Consonants vs. Others & 63.84$\pm$1.45 & 81.21$\pm$0.33 & 80.63$\pm$0.61 & 84.41$\pm$0.92 & 81.54$\pm$0.31 & \textbf{94.03$\pm$0.37}\\
				DS107: Objects vs. Others & 70.17$\pm$0.59 & 76.14$\pm$0.49 & 81.54$\pm$0.92 & 80.92$\pm$0.28 & \textbf{94.23$\pm$0.94} & 92.14$\pm$0.42\\
				DS107: Scrambles vs. Others & 80.73$\pm$0.92 & 77$\pm$1.01 & 85.79$\pm$0.42 & 83.14$\pm$0.47 & 82.23$\pm$0.38 & \textbf{87.05$\pm$0.37}\\
				DS117: Faces vs. Scrambles & 79.36$\pm$0.33 & 83.71$\pm$0.81 & 83.21$\pm$1.23 & 82.29$\pm$0.91 & 94.08$\pm$0.84 & \textbf{94.61$\pm$0.71}\\
				ALL: Faces vs. Others & 61.91$\pm$1.2 & 65.04$\pm$0.99 & 74.9$\pm$0.61 & 78.14$\pm$0.83 & 83.89$\pm$0.28 & \textbf{91.05$\pm$0.12}\\
				ALL: Objects vs. Others & 74.19$\pm$0.92 & 77.88$\pm$0.82 & 73.59$\pm$0.95 & 79.45$\pm$0.77 & 75.61$\pm$0.89 & \textbf{89.24$\pm$0.69}\\
				ALL: Scrambles vs. Others & 79.81$\pm$1.01 & 80$\pm$0.49 & 82.53$\pm$0.83 & 88.14$\pm$0.91 & 88.93$\pm$0.71 & \textbf{92.09$\pm$0.28}\\
				\hline
			\end{tabular}
		\end{small}
	\end{center}
	\vskip -0.25in
\end{table*}
\begin{algorithm}[!t]
\caption{The Proposed Method by using (LOO)}
\label{alg:TheProposedMethod}
\begin{algorithmic}
\STATE {\bfseries Input:} fMRI time series $\mathbf{F}_u, u = 1\text{:}U$, Onsets $\mathbf{S}_u, u=1\text{:}U$, HRF signal $\mathbf{H}$, Gaussian Parameter $\sigma_G$ (default $\sigma_G = 1$):\\
\STATE {\bfseries Output:} MVP performance ($ACC,AUC$)\\
\STATE {\bfseries Method:}\\
1. \textbf{Foreach} Subject $\mathbf{F}_u:$\\ 
2. Create train set $\mathbf{F}_{Tr} = \{\mathbf{F}_j | j=1\text{:}U, j\neq u\}$.\\
3. Extract snapshots of $\mathbf{F}_{Tr}$ by using Algorithm \ref{alg:SnapshotSelection}.\\
4. Generate features of $\mathbf{F}_{Tr}$ by using Algorithm \ref{alg:FeatureExtraction}.\\
5. Train binary classifiers $\eta$ by using $\mathbf{F}_{Tr}$ and \eqref{eq:L1SVM}.\\
6. Generate final predictor ($\eta_{final}$) by using Bagging.\\
7. Consider $\mathbf{F}_u$ as test set.\\
8. Extract snapshots for $\mathbf{F}_u$ by using Algorithm \ref{alg:SnapshotSelection}.\\
9. Generate features for $\mathbf{F}_u$ by using Algorithm \ref{alg:FeatureExtraction}.\\
10. Apply test set on the final predictor ($\eta_{final}$).\\
11. Calculate performance of $\mathbf{F}_u$ ($ACC_i,AUC_i$) \cite{murphy12}.\\ 
12. \textbf{End foreach}\\
13. Accuracy: \cite{murphy12}: $ACC = \sum_{i=1}^{U}ACC_i \big/ U$.\\
14. AUC \cite{murphy12}: $ AUC = \sum_{i=1}^{U}AUC_i \big/ U$.\\
\end{algorithmic}
\end{algorithm}
\vskip -0.25in
\begin{figure}[!h]
	\begin{center}
		\begin{minipage}{0.90\linewidth}
			\includegraphics[width=0.99\textwidth]{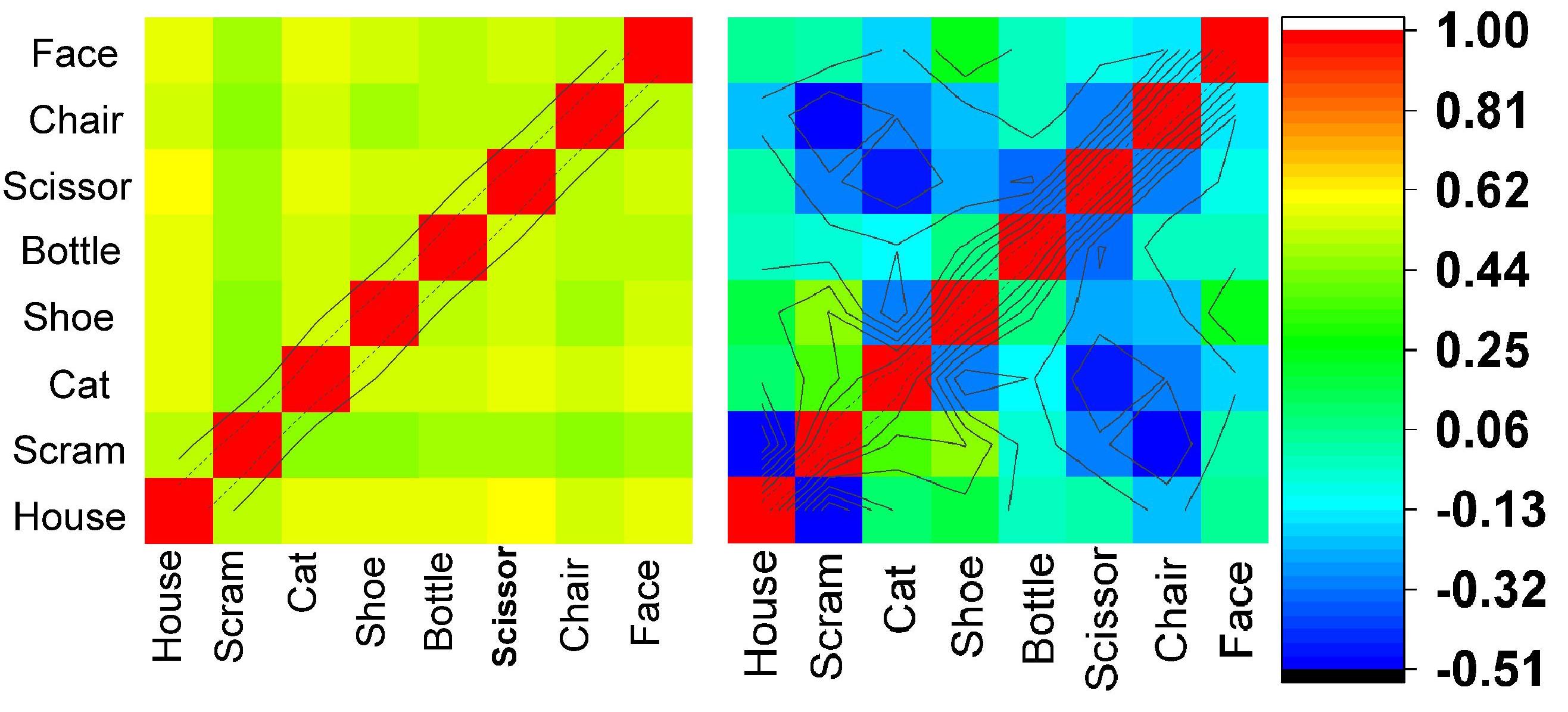}
			\\	\centering (A) \qquad\qquad (B)\\
		\end{minipage}
		\begin{minipage}{0.48\linewidth}
			\includegraphics[width=0.99\textwidth]{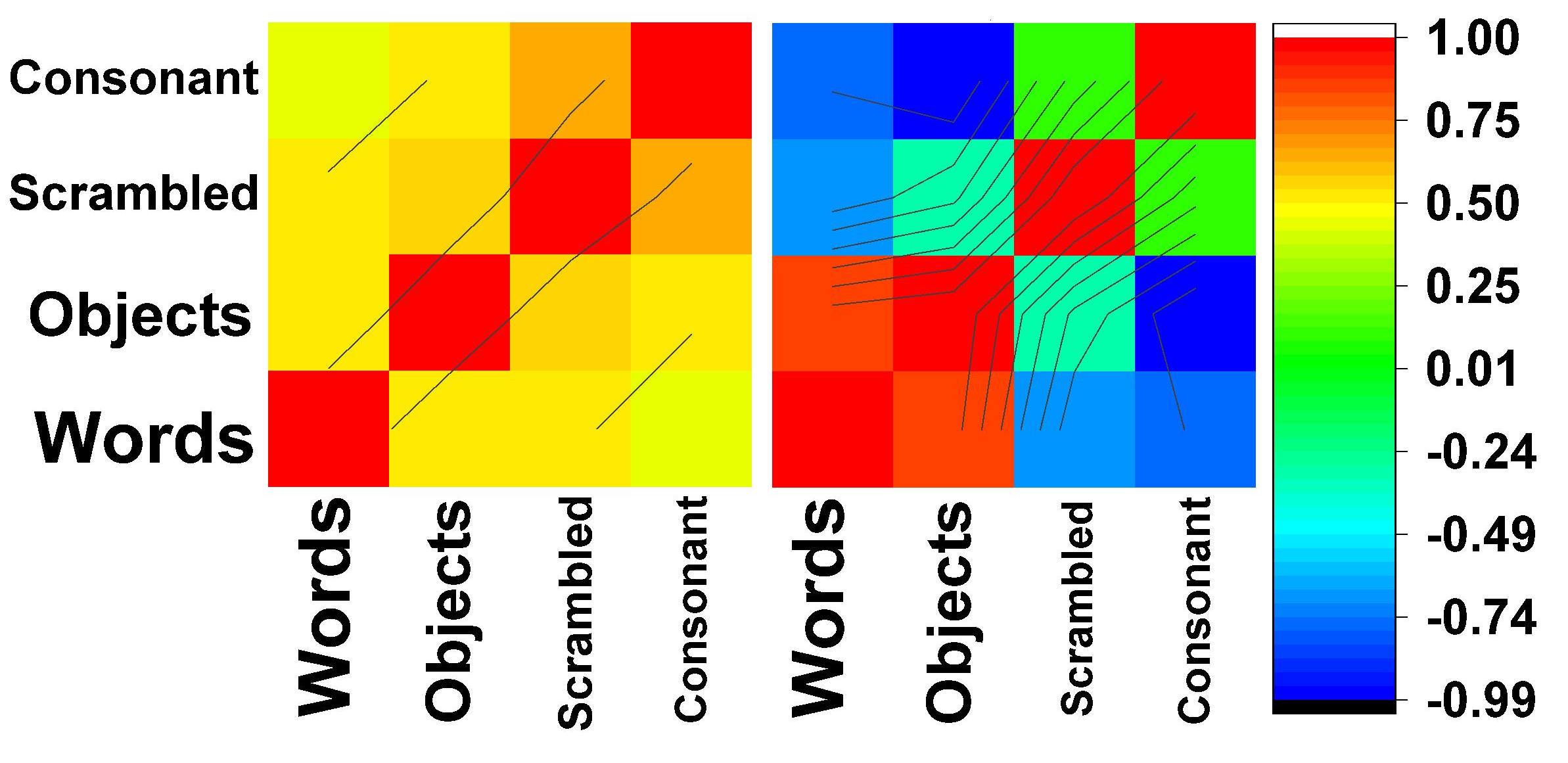}
			\\	\centering (C) \quad (D)
		\end{minipage}		
		\begin{minipage}{0.48\linewidth}
			\includegraphics[width=0.99\textwidth]{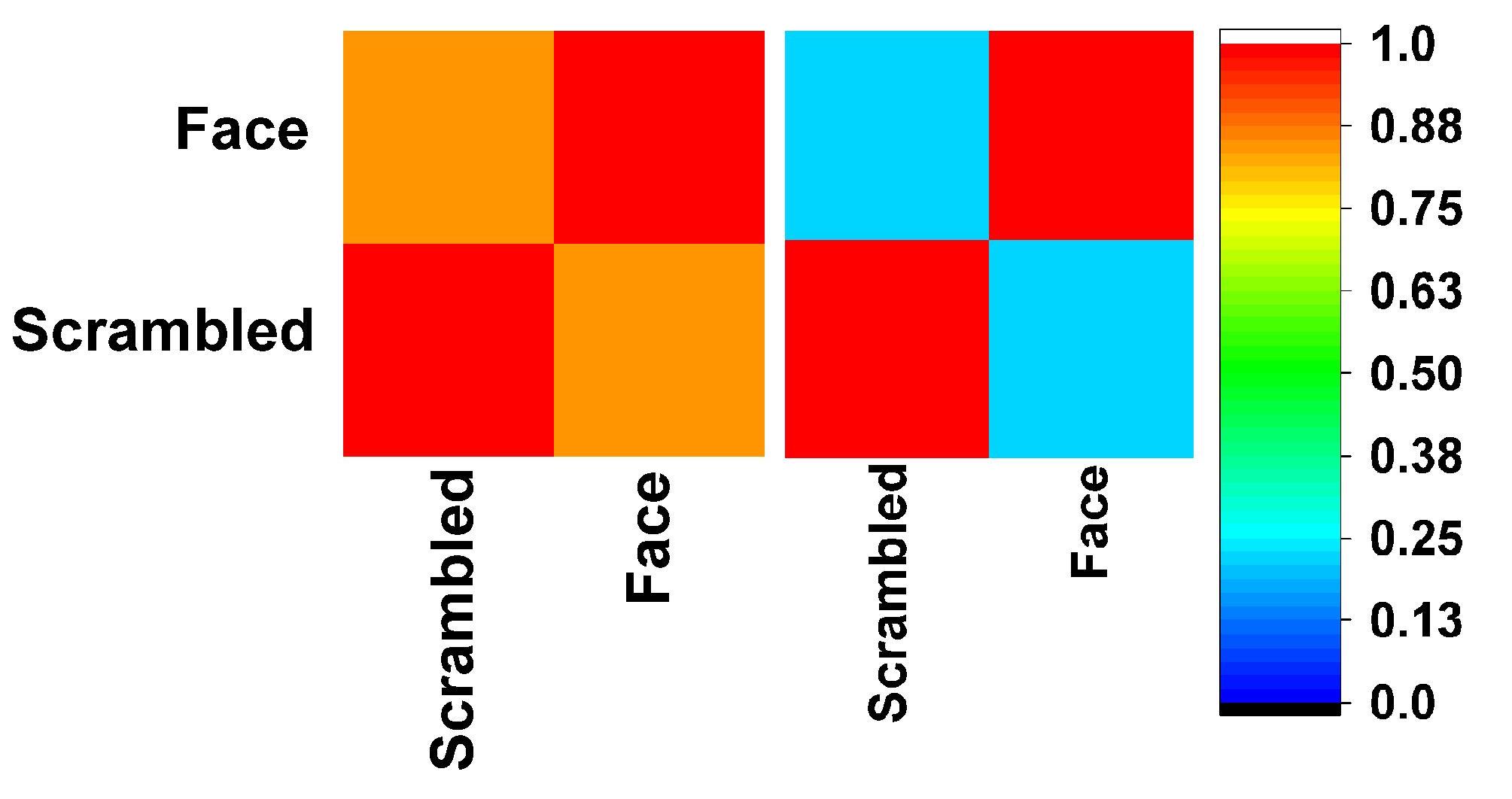}
			\\	\centering (E) \quad (F)
		\end{minipage}
		\caption{Correlation Matrix: for Visual Object Recognition (DS105) data set (A) in the voxel level, (B) feature level, for Word and Object Processing (DS107) data set (C) in the voxel level, (D) feature level, and for multi-subject, multi-modal human neuroimaging dataset (DS117) (E) in the voxel level, (F) feature level.}
		\label{fig:Correlation}
	\end{center}
	\vskip -0.3in
\end{figure}	
\section{Experiments}
\subsection{Data Sets}
This paper utilizes three data sets, shared by openfmri.org, for running empirical studies. As the first data set, `Visual Object Recognition' (DS105) includes $U=71$ subjects. It also contains $p=8$ categories of visual stimuli, i.e. gray-scale images of faces, houses, cats, bottles, scissors, shoes, chairs, and scrambles (nonsense patterns). This data set is analyzed in high-level visual stimuli as the binary predictor, by considering all categories except nonsense photos (scramble) as objects. Please see \cite{haxby14,hanson04,haxby01,haxby12,connolly12} for more information. As the second data set, `Word and Object Processing' (DS107) includes $U=98$ subjects. It contains $p=4$ categories of visual stimuli, i.e. words, objects, scrambles, consonants. Please see \cite{duncan09} for more information. As the last data set, `Multi-subject, multi-modal human neuroimaging dataset' (DS117) includes MEG and fMRI images for $U=171$ subjects. This paper just uses the fMRI images of this data set. It also contains $p=2$ categories of visual stimuli, i.e. human faces, and scrambles. Please see \cite{wakeman15} for more information. These data sets are separately preprocessed by SPM 12 (6685) (www.fil.ion.ucl.ac.uk/spm/), i.e. slice timing, realignment, normalization, smoothing. This paper employs the Montreal Neurological Institute (MNI) 152 T1 1mm as the reference image ($\mathbf{Ref}$) in \eqref{eq:ImageReg} for mapping the extracted snapshots to the standard space ($\widehat{\psi}_i \to \psi_i$). The size of this image in 3D scale is $X=182,Y=218,Z=182$. Moreover,  the \textit{Talairach} Atlas \cite{talairach88} (including $L=1105$ regions) in the standard space is used in \eqref{AutomaticallyDetectedActiveRegions} for extracting features. Further, all of algorithms are implemented in the MATLAB R2016b (9.1) on a PC with certain specifications\footnote{DEL , CPU = Intel Xeon E5-2630 v3 (8$\times$2.4 GHz), RAM = 64GB, OS = Elementary OS 0.4 Loki} by authors in order to generate experimental results.
\subsection{Correlation Analysis}
Figure \ref{fig:Correlation} A, C, and E respectively demonstrate correlation matrix at the voxel level for the data sets DS105, DS107, and DS117. Further, Figure \ref{fig:Correlation} B, D, and F respectively illustrate the correlation matrix in the feature level for the data sets DS105, DS107, and DS117. Since neural activities are sparse, high-dimensional and noisy in voxel level, it is so hard to discriminate between different categories in Figure \ref{fig:Correlation} A, C, and E. By contrast, Figure \ref{fig:Correlation} B, D, and F provide distinctive and informative representation, when the proposed method used the extracted features. 
\subsection {Performance Analysis}
The performance of our proposed method is compared with state-of-the-art algorithms, which were proposed for decoding the visual stimuli in the human brain. As a pioneer algorithm, our method is compared by SVM method \cite{cortes95}, which is used in \cite{hanson04,haxby01} for decoding the visual stimuli. The performance of Graph Net \cite{grosenick13} and Elastic Net \cite{zou05} are reported as the most popular methods for fMRI analysis \cite{mohr15,mcmenamin15,yamashita08,ryali10,carroll09,richiardi11}. Moreover, the performance of L1-Reg. SVM \cite{bradley98} is compared by the proposed method. The L1-Reg. SVM is recently employed by \cite{mohr15} as the most effective approach for decoding visual stimuli. Since this paper also applies L1-Reg. SVM for generating the predictive model in the level of ROIs, it can be considered as a baseline for comparing our feature space with the previous approaches. Lastly, Osher et al. \cite{osher15} proposed a graph-based approach for creating predictors. Indeed, they employed the anatomical structure of the human brain for constructing graph networks. This paper compares the performance of the mentioned methods as well as the proposed method by using LOO cross-validation at the subject level. Further, the Gaussian parameter for smoothing the design matrix is considered $\sigma_G = 1$. The effect of different values of this parameter on the performance of the proposed method will be discussed in the supplementary materials.

Table \ref{tbl:BinaryAccuracy} and \ref{tbl:BinaryAUC} respectively demonstrate the classification Accuracy and Area Under the ROC Curve (AUC) in percentage (\%) for the binary predictors. These tables report the performance of binary predictors based on the category of the visual stimuli. All visual stimuli in the data set DS105 except nonsense photos (scramble) are considered as the object category for generating these experimental results. In addition, different categories of visual stimuli (including words, consonants, objects, and scrambles) in the DS107 are compared by using one-versus-all strategy. Moreover, face recognition based on neural activities is trained by using DS117 data set. Finally, all data sets are combined for generating predictive models for different categories of visual stimuli, i.e. faces, objects, and scrambles. As Table \ref{tbl:BinaryAccuracy} and \ref{tbl:BinaryAUC} demonstrate, the proposed algorithm has generated better performance in comparison with other methods because it provided a better representation of neural activities by exploiting the snapshots of the automatically detected active regions in the human brain. The last three rows in Table \ref{tbl:BinaryAccuracy} and \ref{tbl:BinaryAUC} illustrate the accuracy of the proposed method by combining all data sets. As depicted in these rows, the performances of other methods are significantly decreased. As mentioned before, it is the normalization problem. In addition, our framework employs the extracted features from the automatically detected snapshots instead of using all or a group of voxels, which can decrease noise and sparsity and remove high-dimensionality. Therefore, the proposed method can significantly decrease the time and space complexities and increase rapidly the performance and robustness of the predictive models. 	
\section{Conclusion}
As a conjunction between neuroscience and computer science, Multivariate Pattern (MVP) is mostly used for analyzing task-based fMRI data set. There is a wide range of challenges in the MVP techniques, i.e. decreasing noise and sparsity, defining effective regions of interest (ROIs), visualizing results, and the cost of brain studies. In overcoming these challenges, this paper proposes Multi-Region Neural Representation as a novel feature space for decoding visual stimuli in the human brain. The proposed method is applied in three stages: firstly, snapshots of brain image (each snapshot represents neural activities for a unique stimulus) are selected by finding local maximums in the smoothed version of the design matrix. Then, features are generated in three steps, including normalizing to standard space, segmenting the snapshots in the form of automatically detected anatomical regions, and removing noise by Gaussian smoothing in the level of ROIs. Experimental studies on 4 visual categories (words, objects, consonants and nonsense photos) clearly show the superiority of our proposed method in comparison with state-of-the-art methods. In addition, the time complexity of the proposed method is naturally lower than the previous methods because it employs a snapshot of brain image for each stimulus rather than using the whole of time series. In future, we plan to apply the proposed method to different brain tasks such as risk, emotion and etc.
\section*{Acknowledgment}
We thank the anonymous reviewers for comments. This work was supported in part by the National Natural Science Foundation of China (61422204 and 61473149), Jiangsu Natural Science Foundation (BK20130034) and NUAA Fundamental Research Funds (NE2013105).


\begin{thebibliography}{99}
\bibitem{anderson10} M.~L. Anderson and T. Oates, {\em A critique of multi-voxel pattern analysis}, Proceedings of the 32nd Annual Meeting of the Cognitive Science Society, 2010, pp.~1511--16.
\bibitem{lorbert12} A. Lorbert and P.~J. Ramadge, {\em Kernel hyperalignment}, Advances in Neural Information Processing Systems, 2012, pp.~1790--1798.
\bibitem{mohr15} H. Mohr, U. Wolfensteller, S. Frimmel and H. Ruge, {\em Sparse regularization techniques provide novel insights into outcome integration processes}, NeuroImage, Elsevier, 104 (2015), pp.~163--176.
\bibitem{mcmenamin15} B.~W. McMenamin, R.~G. Deason, V.~R. Steele, W. Koutstaal and C.~J. Marsolek, {\em Separability of abstract-category and specific-exemplar visual object subsystems: Evidence from fMRI pattern analysis}, Brain and Cognition, Elsevier, 93 (2015), pp.~54--63.
\bibitem{haxby14} J.~V. Haxby, A.~C. Connolly, J.~S. Guntupalli, {\em Decoding neural representational spaces using multivariate pattern analysis}, Annual Review of Neuroscience, Annual Reviews, 37 (2014), pp.~435--456.
\bibitem{hanson04} S.~J. Hanson, T.~ Matsuka and J.~V. Haxby, {\em Combinatorial codes in ventral temporal lobe for object recognition: Haxby (2001) revisited: is there a `face' area?}, NeuroImage, Elsevier, 23 (2004), pp.~156--166.
\bibitem{chen15} P.~H.~C. Chen, J. Chen, Y. Yeshurun, U. Hasson, J.~V. Haxby and P.~J. Ramadge, {\em A Reduced-Dimension fMRI Shared Response Model}, Advances in Neural Information Processing Systems, 2015, pp.~460--468.
\bibitem{bradley98} P.~S. Bradley and O.~L. Mangasarian, {\em Feature selection via concave minimization and support vector machines}, International Conference on Machine Learning (ICML), 98 (1998), pp.~82--90.
\bibitem{breiman96} L. Breiman, {\em Bagging predictors}, Machine learning, Springer, 24 (1996), pp.~123--140.
\bibitem{murphy12} K.~P. Murphy, {\em Machine learning: a probabilistic perspective}, MIT press, 2012.
\bibitem{haxby01} J.~V. Haxby, M.~I. Gobbini,  M.~L. Furey, A. Ishai, J.~L. Schouten and P. Pietrini, {\em Distributed and overlapping representations of faces and objects in ventral temporal cortex}, Science, American Association for the Advancement of Science, 293 (2001), pp.~2425--2430.
\bibitem{osher15} D.~E. Osher, R.~R. Saxe, K. Koldewyn, J.~D.~E. Gabrieli, N. Kanwisher, and Z.~M. Saygin, Zeynep M, {\em Structural connectivity fingerprints predict cortical selectivity for multiple visual categories across cortex}, Cerebral Cortex, Oxford University Press, 2003, pp. bhu303. 
\bibitem{formisano08} E. Formisano, F. De Martino, M. Bonte and R. Goebel,{\em `Who' Is Saying `What'? Brain-Based Decoding of Human Voice and Speech}, Science, American Association for the Advancement of Science, (322) 2008, pp.~970--973. 
\bibitem{haxby12} J.~V. Haxby, {\em Multivariate pattern analysis of fMRI: the early beginnings}, NeuroImage, Elsevier, 62 (2012), pp.~852--855.
\bibitem{norman06} K.~A. Norman, S.~M. Polyn, G.~J. Detre and J.~V. Haxby, {\em Beyond mind-reading: multi-voxel pattern analysis of fMRI data}, Trends in Cognitive Sciences, Elsevier, 10 (2006), pp.~424--430.
\bibitem{yamashita08} O. Yamashita, M.~A. Sato, T. Yoshioka, F. Tong and Y. Kamitani, {\em Sparse estimation automatically selects voxels relevant for the decoding of fMRI activity patterns}, NeuroImage, Elsevier, 42 (2008), pp.~ 1414--1429.
\bibitem{ryali10} S. Ryali, K. Supekar, D.~A. Abrams and V. Menon, {\em Sparse logistic regression for whole-brain classification of fMRI data}, NeuroImage, Elsevier, 51 (2010), pp.~752--764.
\bibitem{zou05}	H. Zou and T. Hastie, {\em Regularization and variable selection via the elastic net}, Journal of the Royal Statistical Society: Series B (Statistical Methodology), Wiley Online Library, 67 (2005), pp.~301--320.
\bibitem{carroll09} M.~K. Carroll, G.~A. Cecchi, I. Rish, R. Garg and A.~R. Ravishankar, {\em Prediction and interpretation of distributed neural activity with sparse models}, NeuroImage, Elsevier, 44 (2009), pp.~112--122.
\bibitem{richiardi11} J. Richiardi, H. Eryilmaz, S. Schwartz, P. Vuilleumier and D. Van De Ville, {\em Decoding brain states from fMRI connectivity graphs}, Neuroimage, Elsevier, 56 (2011), pp.~616--626.
\bibitem{varoquaux12} G. Varoquaux, A. Gramfort and B. Thirion, {\em Small-sample brain mapping: sparse recovery on spatially correlated designs with randomization and clustering}, International Conference on Machine Learning (ICML), 2012.
\bibitem{cortes95} C. Cortes and V. Vapnik, {\em Support-vector networks}, Machine learning, Springer, 20 (1995), pp.~273--297.
\bibitem{grosenick13} L. Grosenick, B. Klingenberg, K. Katovich, B. Knutson and J.~E. Taylor, {\em Interpretable whole-brain prediction analysis with GraphNet}, NeuroImage, Elsevier, 72 (2013), pp.~304--321.
\bibitem{friston03} K.~J. Fristo,  and J.~O.~H.~N.~ Ashburner  and J.~ Heather and others, {\em Statistical parametric mapping}, Neuroscience Databases: A Practical Guide, 2003, pp.~237.
\bibitem{jenkinson02} M. Jenkinson, P. Bannister, M. Brady and S. Smith, {\em Improved optimization for the robust and accurate linear registration and motion correction of brain images}, NeuroImage, Elsevier, 17 (2002), pp.~825--841.
\bibitem{connolly12} A.~C. Connolly, J.~S. Guntupalli, J. Gors, M. Hanke, Y.~O. Halchenko, Y.~C. Wu, H.~ Abdi and J.~V. Haxby, {\em The representation of biological classes in the human brain}, The Journal of Neuroscience, 32 (2012), pp.~2608--2618.
\bibitem{duncan09} K.~J. Duncan, C. Pattamadilok, I. Knierim and T.~J. Devlin, {\em Consistency and variability in functional localisers}, NeuroImage, Elsevier, 46 (2009), pp.~1018--1026.
\bibitem{wakeman15} D.~G. Wakeman and R.~N. Henson, {\em A multi-subject, multi-modal human neuroimaging dataset}, Scientific data, Nature Publishing Group, 2 (2015).
\bibitem{talairach88} J. Talairach and P. Tournoux, {\em Co-planar stereotaxic atlas of the human brain. 3-Dimensional proportional system: an approach to cerebral imaging}, Thieme, 1988.
\end{thebibliography}
\end{document}